\documentclass[peerreview]{ieeecolor}
\pdfoutput=1
\usepackage{jsen}
\usepackage{cite}
\usepackage{amsmath,amssymb,amsfonts}
\usepackage{algorithmic}
\usepackage{graphicx}
\usepackage{textcomp}
\usepackage{wrapfig}
\usepackage{caption}
\usepackage{subcaption}
\usepackage{tabularray}
\usepackage{tabularx}

\graphicspath{ {./Figures/} }

\def\BibTeX{{\rm B\kern-.05em{\sc i\kern-.025em b}\kern-.08em
    T\kern-.1667em\lower.7ex\hbox{E}\kern-.125emX}}
\definecolor{abstractbg}{rgb}{0.89804,0.94510,0.83137}
\begin{document}
\title{License Plate Super-Resolution Using Diffusion Models} 
\author{Sawsan~AlHalawani, \IEEEmembership{Member, IEEE}, Bilel Benjdira, \IEEEmembership{Senior, IEEE}, Adel Ammar, Anis Koubaa, \IEEEmembership{Senior, IEEE},  Anas~M.~Ali}


\IEEEtitleabstractindextext{%
\fcolorbox{abstractbg}{abstractbg}{%
\begin{minipage}{\textwidth}%
\begin{abstract}
In surveillance, accurately recognizing license plates is hindered by their often low quality and small dimensions, compromising recognition precision. Despite advancements in AI-based image super-resolution, methods like Convolutional Neural Networks (CNNs) and Generative Adversarial Networks (GANs) still fall short in enhancing license plate images. This study leverages the cutting-edge diffusion model, which has consistently outperformed other deep learning techniques in image restoration. By training this model using a curated dataset of Saudi license plates, both in low and high resolutions, we discovered the diffusion model's superior efficacy. The method achieves a 12.55\% and 37.32\% improvement in Peak Signal-to-Noise Ratio (PSNR) over SwinIR and ESRGAN, respectively. Moreover, our method surpasses these techniques in terms of Structural Similarity Index (SSIM), registering a 4.89\% and 17.66\% improvement over SwinIR and ESRGAN, respectively. Furthermore, 92\% of human evaluators preferred our images over those from other algorithms. In essence, this research presents a pioneering solution for license plate super-resolution, with tangible potential for surveillance systems.

\end{abstract}
\thanks{The authors are with the Robotics and Internet of Things Lab (RIOTU Lab)  at Prince Sultan University, Riyadh, 11586, Saudi Arabia (email: shalawani@psu.edu.sa, bbenjdira@psu.edu.sa, aammar@psu.edu.sa, akoubaa@psu.edu.sa, aaboessa@psu.edu.sa). }
\end{minipage}}}

\maketitle


\section{Introduction}
\label{sec:introduction}

\IEEEPARstart{A}{s} the number of motor vehicles continues to rise, license plates become crucial to public security \cite{8643795}. Computer vision techniques offer timely solutions for license plate identification. A prevalent technique is License Plate Recognition (LPR) \cite{Zhuang2018}, which transforms license plate images into recognizable character sets. This aids in tasks such as vehicle tracking for security, traffic violations monitoring \cite{9226453}, toll collection \cite{7922567}, and intelligent parking \cite{8471111}.

The effectiveness of LPR hinges on the quality of its input images. While surveillance cameras capture these images, many output images of subpar quality and low resolution. Factors such as camera positioning, weather, and lighting conditions play roles in this degradation. Given that license plates occupy a minor section of these images, they often appear at even lower resolutions. Enhancing and denoising these images is vital for practical applications and remains an active research area.

Super Resolution (SR) is a reconstruction technology that produces high-resolution images from low-resolution images. Super-resolution has gained researchers attention recently. 
The first paper that addressed this problem was by Dong et al. \cite{Dong2015FirstSR}. They defined the mapping between low and high-resolution images using a Convolutional Neural Network (CNN).  They demonstrated good restoration quality that was fast to achieve. Since then, many research studies have been published in the area with better results. 

\begin{figure}
     \centering
     \begin{subfigure}[b]{0.45\columnwidth}
         \centering
         \includegraphics[width=\columnwidth]{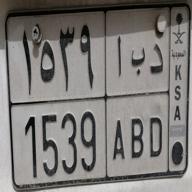}
         \caption{Ground truth}
         \label{}
     \end{subfigure}
     \hfill
     \begin{subfigure}[b]{0.45\columnwidth}
         \centering
         \includegraphics[width=\columnwidth]{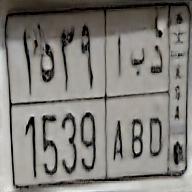}
         \caption{ESRGAN output}
         \label{}
     \end{subfigure}
     \hfill
     
     

     

     
     \begin{subfigure}[b]{0.45\columnwidth}
         \centering
         \includegraphics[width=\columnwidth]{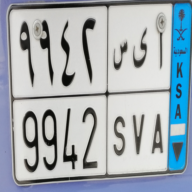}
         \caption{Ground truth}
         \label{}
     \end{subfigure}
     \hfill
     \begin{subfigure}[b]{0.45\columnwidth}
         \centering
         \includegraphics[width=\columnwidth]{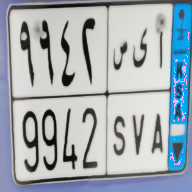}
         \caption{ SwinIR output}
         \label{}
     \end{subfigure}
        \caption{ Comparison between the ground truth and the super resolved images using SwinIR and ESRGAN. } 
        \label{fig:FailureCasesIntro}
\end{figure}

There has been some research work that addressed enhancing LP images using SR techniques. 
Some authors used older methods such as image processing and interpolation techniques \cite{Ghoneim2017}. Such approaches depend on the pixel information, which may lead to poor quality.
Another approach  utilized multiple low resolution images to produce a single high resolution image \cite{GUARNIERI2021}.  Other approaches \cite{Zhang2018,Lee2020, HAMDI2021} utilized SRGAN model to produce the high quality images. 
Figure \ref{fig:FailureCasesIntro} shows some examples where the traditional methods could fail to produce images which mimic exactly the original image with high quality.

Recently, the Denoising Diffusion Probabilistic Model (DDPM) \cite{Diffusion2015} has garnered significant interest, often surpassing alternatives like GAN-based methods. DDPM's superior performance is evident across diverse domains, such as image generation \cite{Dhariwal2021}, super-resolution \cite{Saharia2022, Lin2021}, deblurring \cite{Whang2021}, and more. Notably, while DDPM predominantly addresses general images, its application for license plate super-resolution remains unexplored.

In this study, we pioneer the adaptation of Diffusion Models for License Plate (LP) image Super Resolution. Our dataset comprises genuine Saudi Arabia license plate images. Through comprehensive experiments, we benchmarked our approach against leading Super Resolution methods like SwinIR \cite{SwinIR2021} and ESRGAN \cite{ESRGAN2018}, employing PSNR, SSIM, and MS-SSIM as evaluation metrics. Notably, our diffusion model enhanced PSNR values by 13\% and 37\% over SwinIR and ESRGAN respectively, with SSIM improving by 5\% and 18\%. A human-centric evaluation confirmed our method's efficacy: 92\% of participants favored our generated images over those of SwinIR and ESRGAN. This underscores the potential of diffusion models in reconstructing high-fidelity LP images, indicating an exciting avenue for future research.

The primary contributions of this study are: 
\begin{itemize}
  \item Introduction of a novel method for License Plate Image Super Resolution, demonstrating superior performance over existing state-of-the-art approaches. 
  \item Pioneering the application of Diffusion Models for license plate image super resolution, to the best of our knowledge.
  \item Development of a unique dataset focused on Saudi Arabian License Plates. 
\end{itemize}

Section \ref{related-work} presents a summary of related work in license plate image SR. Our methodology is elucidated in Section \ref{method}. Comprehensive experimental results, both quantitative and qualitative, are discussed in Section \ref{experiments}. The paper concludes with discussions in Section \ref{Discussion} and final remarks in Section \ref{Conclusions}.

\section{Related Works} \label{related-work}

This section reviews prior research pertinent to License Plate Image Super Resolution.

Image Super Resolution (ISR) aims to produce high-resolution images from their lower-resolution counterparts. Although ISR has wide applications, this paper focuses on license plates. Recent methodologies predominantly employ deep learning, especially convolution neural networks (CNNs) and Generative Adversarial Networks (GANs).

Yang et al. \cite{Yang2018} introduced a multi-scale super-resolution CNN (MSRCNN), drawing inspiration from GoogLeNet \cite{szegedy2014going}. Their model has three convolutional layers designed for feature extraction, feature mapping, and high-resolution image reconstruction. Importantly, they found that training with license plate datasets results in superior super-resolution specifically for license plate images.

Lee et al. \cite{Lee2020} built upon the SRGAN architecture and incorporated a character-based perceptual loss for high-quality image generation suitable for Optical Character Recognition (OCR). Their method, when compared with the traditional SRGAN, improved OCR accuracy by 6.8\% and 4.0\% at the plate and character levels, respectively.

Hamdi et al. \cite{HAMDI2021} developed a Double GAN for image enhancement and super-resolution. While their model exhibited significant image quality improvement, it was computationally intensive.

Lin et al. \cite{Lin2021} targeted Chinese License Plate image enhancement using a GAN. They introduced a residual dense network and progressive upsampling, achieving better performance metrics and reduced reconstruction time compared to several benchmarks.

Shah et al. \cite{Shah2022} employed the VGG-19 network in conjunction with a GAN for super-resolution, effectively addressing the over-smoothing issue. Their approach demonstrated a marked improvement in PSNR and accuracy.

Nascimento et al. \cite{Nascimento2022} extended the Multi-Path Residual Network (MPRNet) \cite{mehri2020mprnet}, merging PixelShuftle layers with attention modules. Their model, designed to enhance License Plate images, outperformed the baseline MPRNet in various aspects.

Lee et al. \cite{Lee2022AlternativeCL} proposed a two-step framework, first training an ISR network, followed by a character recognition network. While their global image extraction technique showed promise, it faltered under challenging conditions like low lighting and rotations.

In another contribution, \cite{Lee2022} combined ISR with feature extraction for License Plate recognition. Their approach optimized weight freezing techniques to leverage high-frequency image details and character information, albeit with extended training durations.

Our research pioneers the adaptation of diffusion models \cite{Diffusion2015} for LP super-resolution. To date, the diffusion model, while utilized for generic image denoising, has not been specifically applied to LP images. Notably, Ho et al. \cite{HO2020} and Saharia et al. \cite{Saharia2022} showcased its potential in denoising and super-resolution for general and remote sensing images, respectively.

A summary of the discussed research can be found in Table \ref{fig:relatedWorkSummary}.

\begin{table*}[]
\caption{A summary of the related work which is addressed in this paper}
\label{fig:relatedWorkSummary}
\begin{tabularx}{\textwidth}{|p{0.3cm}|X|X|p{0.7cm}|p{0.7cm}|p{0.5cm}|X|}

\hline
\bf{Ref} & \bf{Key Contributions} & \bf{LP Dataset} & \bf{PSNR dB} & \bf{SSIM} & \bf{Year} \\
\hline
\cite{Yang2018} & Multi-scale super-resolution CNN (MSRCNN). It was inspired by the Inception architecture of GoogLeNet.& Datasets are collected by video surveillance systems. 
Dataset1 includes 96 LP images. 
Dataset2 includes 986 LP images. 
& 32.1417 & 0.9363 & 2018 \\
\hline
\cite{Lee2020} & Character-based perceptual loss with SRGAN & Created their dataset from public LP datasets \cite{LeeDataset}. They used 6000 training, 750 validation, and 750 test images. & - & - & 2020 \\
\hline
\cite{HAMDI2021} & Double GAN network (D\_GAN\_ESR) & Custom two datasets: The first added a motion blur noise. The second used a CycleGAN to transfer the style to an analogue image style.
& 29.558 & 0.227 & 2021  \\
\hline
 \cite{Lin2021} & Used residual connections with progressive upsampling & Chinese City Parking Dataset (CCPD) \cite{ChineseLPDataset} & 26.08 & 0.77  & 2021 \\
\hline
\cite{Shah2022} & Pre-trained VGG-19 & LP Images & 28.696 & - & 2022 \\
\hline
\cite{Nascimento2022} & Single Image Reconstruction model with MPRNet and attention modules & Their created LP images dataset was released for public at \cite{nascimento2022combining} & 26.4 & 0.89 & 2022  
\\
\hline
\cite{Lee2022AlternativeCL} & SR network with character recognition network using global image extraction technique & 11,428 images in the training set and 1,999 images in the validation set. & 34.13 & - & 2022 \\
\hline
\cite{Lee2022} & Weight freezing technique, high-frequency feature extraction, global image extraction technique & UFPR and Greek vehicle datasets & 20.6 & - & 2022  \\
\hline
\cite{HO2020}  & diffusion probabilistic models and denoising score matching with Langevin dynamics & CIFAR10 dataset & - & - & 2020 \\
\hline
\cite{Saharia2022}  & Adapts denoising diffusion probabilistic models via repeated refinements  to image-to-image translation & Training face SR on Flickr-Faces-HQ (FFHQ). Evaluating on CelebA-HQ [12]. Training natural image SR on ImageNet 1K. 
& 23.04 & 0.65 & 2022 \\
\hline
\cite{Liu2022} & Proposed  the generative Diffusion Model with Detail Complement (DMDC) into Remote Sensing Super Resolution task & Potsdam (Germany) dataset and Vaihingen (Germany) dataset for remote sensing & 23.46 & 0.6696 & 2022 \\
\hline
\end{tabularx}
\end{table*}

\section{Proposed Method} \label{method}

In this section, we introduce our method dedicated to the super-resolution of license plate images.

Image super-resolution aims to generate high-resolution (HR) images from their low-resolution (LR) counterparts. Such a dataset consists of paired LR and HR images, represented as:

\begin{equation}
D = \{x_i^{lr}, z_i^{hr}\}_{i=1}^N
\end{equation}

Here, \(x^{lr}\) is the degraded LR image with \(z^{hr} \) as its original HR version. The degradation typically stems from introducing noise to the HR image or reducing its resolution, as described by:

\begin{equation}
x^{lr} = G( z^{hr}; \theta)
\end{equation}

where \(\theta\) denotes degradation parameters. The super-resolution task seeks to reconstruct \(y^{hr}\), an approximation of the HR image \(z^{hr} \) from \(x^{lr}\), using the model:

\begin{equation}
y^{hr} = F( x^{lr}; \Phi)
\end{equation}

While many current techniques rely on pixel-level information, this often sacrifices important image characteristics. Recognizing the effectiveness of the diffusion model in image generation, given its proficiency in capturing global features, we employ it for license plate super-resolution.

\begin{figure*}[h] 
\centering
\includegraphics[width=0.95\linewidth]{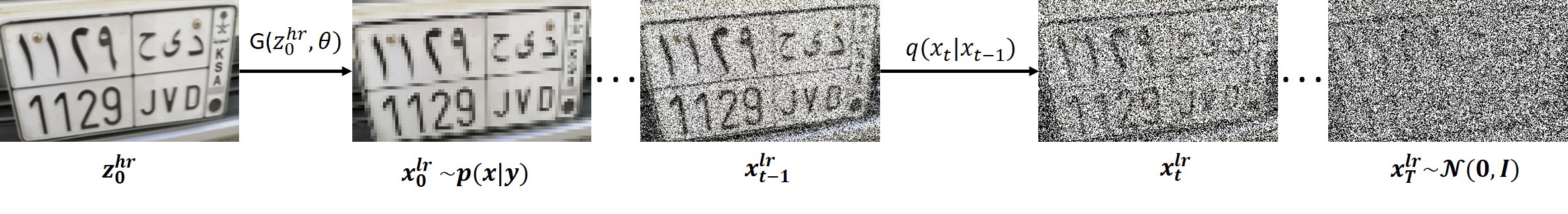}
\caption{Starting from an HR license plate image \(z_0^{hr}\), it's either noise-introduced or downscaled to produce \(x_0^{lr}\). The forward process further adds noise until reaching isotropy at step \(T\).}
\label{fig:ForwardStep}
\end{figure*}

Diffusion models, rooted in Non-Equilibrium Thermodynamics, comprise two phases: the forward noise introduction and the backward noise removal processes. The former uses a Markov chain approach to gradually degrade the image by adding Gaussian noise, visualized in Figure \ref{fig:ForwardStep}. 

From a given input \(x_0^{lr}\), each Markov chain step introduces spherical Gaussian noise with variance \(\beta_t\). At step \(t\), \(X_t^{lr}\) is formed by adding noise to \(X_{t-1}^{lr}\) following:

\begin{equation}\label{eq:forwardNormalDist}
q(x_t^{lr}|x_{t-1}^{lr})=\mathcal{N}(x_t^{lr};\mu_t=\sqrt{1-\beta_t}x_{t-1}^{lr},\sum_t=\beta I)
\end{equation}

This chain produces a sequence from \(x_0^{lr}\) to \(x_T^{lr}\) defined by:

\begin{equation}\label{eq:forwardTrajectory}
q(x_{1:T}^{lr}|x_{0}^{lr})=\prod_{t=1}^T q(x_t^{lr}|x_{t-1}^{lr})
\end{equation}

Using reparametrization, we can succinctly describe \(x_t^{lr}\) without multiple iterations:

\begin{equation}
\begin{aligned}\label{eq:parameterization}
x_t^{lr} &= \sqrt{1-\beta_t}x_{t-1}^{lr}+\sqrt{\beta_t}\epsilon_{t-1} \\
&= \sqrt{\alpha_t}x_{t-2}^{lr}+\sqrt{1-\alpha_t}\epsilon_{t-1}\\
&= \dots \\
&= \sqrt{\overline{\alpha}_t}x_0^{lr}+\sqrt{1-\overline{\alpha}_t}\epsilon_0
\end{aligned}
\end{equation}

Here, \(\alpha_t\) and \(\overline\alpha_t\) are defined, giving us the ability to determine \(x_t^{lr}\) at any step based on \(\beta_t\). Consequently, \(x_t^{lr}\) is produced by:

\begin{equation}
x_t^{lr} \sim q(x_t^{lr}|x_{0}^{lr})=\mathcal{N}(x_t^{lr};\sqrt{\overline\alpha_t}x_{0}^{lr},(1-\overline\alpha_t) I)
\end{equation}

\begin{figure*}[h] 
\centering
\includegraphics[width=0.8\linewidth]{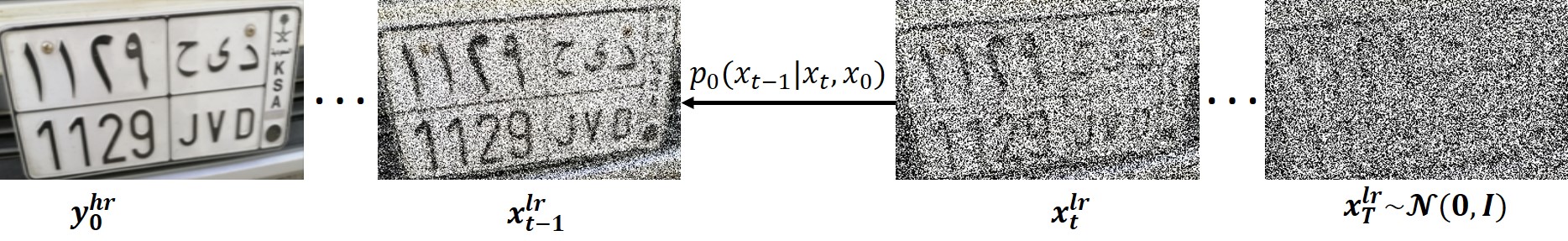}
\caption{From the fully degraded image \(x_T^{lr}\), the backward process reconstructs the HR image \(y_0^{hr}\) using prior knowledge about \(x_0^{lr}\).}
\label{fig:BackwardStep}
\end{figure*}

\textbf{Denoising Process.}
The output of the previous step produces a set of noisy images \(X_t^{lr}\) that has an isotropic distribution. Our target at this stage is to recover the high resolution LP image from the noisy image. This is done by reversing the previous process and gradually removing the noise from the \(x_t^{lr}\) sample which is drawn from the normal distribution \(q(x_0^{lr})\) to obtain a sample from the original date distribution. The backward process is shown in Figure \ref{fig:BackwardStep}. This can be approximated with the parameterized model \(p_0\) that uses a CNN as follows:

\begin{equation}\label{eq:reverseDistribution}
    p_0(x_{t-1}^{lr}|x_t^{lr})=\mathcal{N}(x_{t-1}^{lr};\mu_\theta(x_t^{lr},t),\Sigma_\theta(x_t^{lr}, t))
\end{equation}

This model \(p(x_0^{lr})\) can be used to predict the \(\mu_\theta(x_t^{lr},t)\) and \(\Sigma_\theta(x_t^{lr}, t)\) parameters of the Gaussian distribution at each time-step \(t\). Then, the trajectory from \(x_T^{lr}\) to \(x_0^{lr}\) can be found as follows: 

\begin{equation} \label{eq:reverseTrajectory}
    p_0(x_{0:T}^{lr})=p_0(x_T^{lr})\prod_{t-1}^T p_0(x_{t-1}^{lr}|x_t^{lr})
\end{equation}

In order for the generative model to recover the high LP resolution image, it needs to have some information about the original image \(x_0^{lr}\). Therefore, we can condition the sampling of \(x_t^{lr}\) at time-step \(t\) based on \(x_0^{lr}\). This can be represented as: 

\begin{equation}
    q(x_{t-1}^{lr}|x_t^{lr},x_0^{lr})=\mathcal{N}(x_{t-1}^{lr}; \tilde\mu(x_t^{lr},x_0^{lr}),\tilde\beta_t I)
\end{equation}

\begin{equation}
    \tilde\beta_t = \frac{1-\overline\alpha_{t-1}}{1-\overline\alpha_t}\cdot  \beta_t
\end{equation}

\begin{equation}
\tilde\mu_t(x_t^{lr}, x_0^{lr}) = \frac{\sqrt{\overline\alpha_{t-1}}\beta_t}{1-\overline\alpha_t} x_0^{lr} + \frac{\sqrt{\alpha_t}(1-\overline\alpha_{t-1})}{1-\overline\alpha_t}x_t^{lr}
\end{equation}

Therefore, we can express \(x_0^{lr}\), given that \(\epsilon \sim \mathcal{N}(0,I)\),  as follows:
\begin{equation}
    x_0^{lr} = \frac{1}{\sqrt{\overline\alpha_t}}(x_t^{lr}-\sqrt{1-\overline\alpha_t} \epsilon)
\end{equation}

Besides, we can define a neural network \(\epsilon_\theta(x_t^{lr}, t)\) to find the values of \(\epsilon\) and the mean as follows: 

\begin{equation}
    \tilde\mu_\theta(x_t^{lr}, t)= \frac{1}{\sqrt{\alpha_t}}(x_t^{lr}-\frac{\beta_t}{\sqrt{1-\overline\alpha_t}}\epsilon_\theta(x_t^{lr},t))
\end{equation}

Diffusion models define the loss function as the Mean Square Error (MSE) between the added noise to the original image and the predicted noise in the backward process. We follow the same approach as Ho et al. \cite{HO2020} by using a simplified version of the loss function that ignores the weighting coefficients  and outperforms its full original version. It can be  expressed as follows: 

\begin{equation}
    L_t =\mathbb{E}_{x_0^{lr},t,\epsilon}\left[ ||\epsilon - \epsilon_0(\sqrt{\overline\alpha_t} x_0^{lr}  + \sqrt{1-\overline\alpha_t} \epsilon, t)||^2\right]
\end{equation}

Our model utilizes a U-Net  architecture \cite{u-net2015} which consists of a contracting  and an expanding paths. 
The contracting path is mainly responsible for increasing the image resolution. 
It has  successive layers that capture the context using up-sampling operators.  
The  expanding path  enables precise localization by combining the high resolution features from the contracting path with the up-sampled output. 
This is followed by a convolution layer that learns to generate a more precise image. 
The contracting path uses a large number of feature channels which are used to pass information to the higher resolution layers. Therefore, the expanding path will be symmetric to the contracting path, which produces a U-shaped network architecture rather than a fully connected network.   

%


\section{Experiments} \label{experiments}

\subsection{Datasets} \label{Datasets}
Our dataset consists of genuine images of Saudi Arabian License Plates (LP). These images, captured with cameras, were meticulously cropped to focus solely on the license plates, excluding extraneous background details. Saudi license plates, featuring a mix of Arabic and English characters and numerals, represent various designs, all of which are encompassed in our study. It's essential to note that while these mixed characters might challenge certain techniques, our method sidesteps character recognition, ensuring such variations don't influence our performance.

The dataset comprises 593 color images with dimensions \(192\times192\times3\), each with red (R), green (G), and blue (B) channels. For model development, we allocated 92\% of the images (543 images) for training and reserved the remaining 8\% (50 images) for testing. Original images underwent downsampling by a factor of 4, resulting in low-resolution images of \(48\times48\times3\). These downsampled images were used as inputs to the diffusion model. For validation, our model was tested on downscaled images with identical resolution (\(48\times48\times3\)), emphasizing the versatility and robustness of our approach without further fine-tuning.

\subsection{Evaluation Metrics}\label{EvalMetrics}
In order to assess the quality of our model comprehensively, we used the following metrics: Peak Signal-to-Noise Ratio (PSNR), Structural Similarity (SSIM) \cite{PSNR_SSIM} and Multi-scale Structural Similarity (MS-SSIM) \cite{MSSSIM}. 
PSNR represents  the ratio between the maximum possible value (power) of a signal and the power of the distorting noise that degrades the quality of the image.  
PSNR is well suited in this context due to  its simple calculations and clear meaning. It is based on using the Mean Squared Error (MSE) which compares the original  pixel values  to the degraded image. It can be defined as follows: 

\begin{equation}
    MSE = \frac{\sum_1^{M*N} [x-y]^2}{M*N}
\end{equation}

\begin{equation}\label{eq:PSNR}
    PSNR = 10 * log_{10}\frac{R^2}{MSE}
\end{equation}

where R is defined based on the image format and \(M\) and \(N\) are the number of rows and columns of the image.

SSIM relies on extracting structural information from the image by assuming pixels interdependency which is close to the human perception. SSIM is defined based on three characteristics which are captured from the image, i.e. luminance, contrast and structure. They are defined as follows: 

\begin{equation}\label{eq:luminance}
    luminance(x,y)=\frac{2\mu_x\mu_y + \alpha_1}{\mu^2_x + \mu^2_y + \alpha_1}
\end{equation}

\begin{equation}\label{eq:contrast}
    contrast(x,y)=\frac{2\mu_{xy} + \alpha_2}{ \sigma^2_x + \sigma^2_y+ \alpha_2}
\end{equation}

\begin{equation}\label{eq:structure}
    structure(x,y)=\frac{\sigma_{xy}+ \alpha_3}{ \sigma_x \sigma_y + \alpha_2}
\end{equation}

 where \(\mu_x\) and \(\mu_y\) represent the mean of \(x, y\) respectively. \(\sigma_x\) and \(\sigma_y\) represent the standard deviation of \(x, y\) respectively. \(\sigma_{xy}\) is the covariance of \(x, y\). For division stabilization, \(\alpha_1, \alpha_2 \) and \(\alpha_3\) constants are used. A simplified representation of SSIM can be written as follows: 

 \begin{equation}
     SSIM(x,y)=\frac{ (2\mu_x \mu_y + \alpha_1) (\sigma_{xy} + \alpha_2) }{(\mu^2_x + \mu^2_y + \alpha_1) (\sigma^2_x + \sigma^2_y + \alpha2)  }
 \end{equation}

MS-SSIM performs multi-step downsampling steps which provides more flexibility than SSIM and incorporates the variations of image resolution and viewing conditions. MS-SSIM can be written as follows: 

\begin{equation}
    MS-SSIM(x,y)=l_M(x,y)^{\alpha_M}. \prod_{j=1}^M [c_j(x,y)]^{\beta_j}[s_j(x,y)]^{\gamma_j}
\end{equation}

where \(l_M\) is the luminance that is calculated by eq. \ref{eq:luminance} at scale \(M\) only.   Moreover, \(c_j, s_j \) are the contrast and structure which are calculated based on eq. \ref{eq:contrast} and eq. \ref{eq:structure}, respectively. The constants \(\alpha_M, \beta_j, \gamma_j\) are used to adjust the relative importance of each term.  
\subsection{Implementation Details}

This section delineates the specifics of our diffusion model implementation and provides a comparative overview with the SwinIR and ESRGAN models.

Our diffusion model leveraged Pytorch for its implementation. Conversely, SwinIR utilized Pytorch Lightning, while ESRGAN was built upon TensorFlow. For the computational demands of our diffusion model and SwinIR, we employed the NVIDIA Quadro RTX 8000 (48601 MiB) GPU. ESRGAN, however, was implemented using the NVIDIA Tesla T4 (15360 MiB) provided by Google Colab.

The construction of our diffusion model involved an initial step of downsizing our authentic HR images from a resolution of \(192\times192\times3\) to \(48\times48\times3\). To enhance the model's robustness, we augmented the dataset through random rotations at angles of [5, 10, 15]. The training regimen consisted of 8688 iterations spanning 64 epochs, with each batch comprising 4 images. Drawing from Saharia et al. \cite{Saharia2022}, we adopted a fixed learning rate of 0.0002, incorporating a linear warm-up schedule via the Adam Optimizer.

Table \ref{TrainingTime} indicates that the diffusion model incurs a 21\% longer training duration compared to SwinIR and 2.6\% relative to ESRGAN. Nonetheless, as the ensuing section will demonstrate, the resultant performance more than compensates for this marginal increase in time.

\begin{table} 
\caption{This table shows the training time taken by each of the methods to train the LP super resolution model.}
\label{TrainingTime}
\centering
\setlength{\tabcolsep}{3pt}
\begin{tblr}{|c|c|c|c|c|}
\hline
 & \textbf{SwinIR}	& \textbf{ESRGAN}	& \textbf{Diffusion Model}\\
\hline
Training Time (seconds)		& 9676			& 11404      & 11700 \\ 
\hline
\end{tblr}

\end{table}


\section{Discussion} \label{Discussion} 

\subsection{Visualization of the Reconstruction Steps}
\begin{figure}[h]
\centering
\includegraphics[width=0.9\columnwidth]{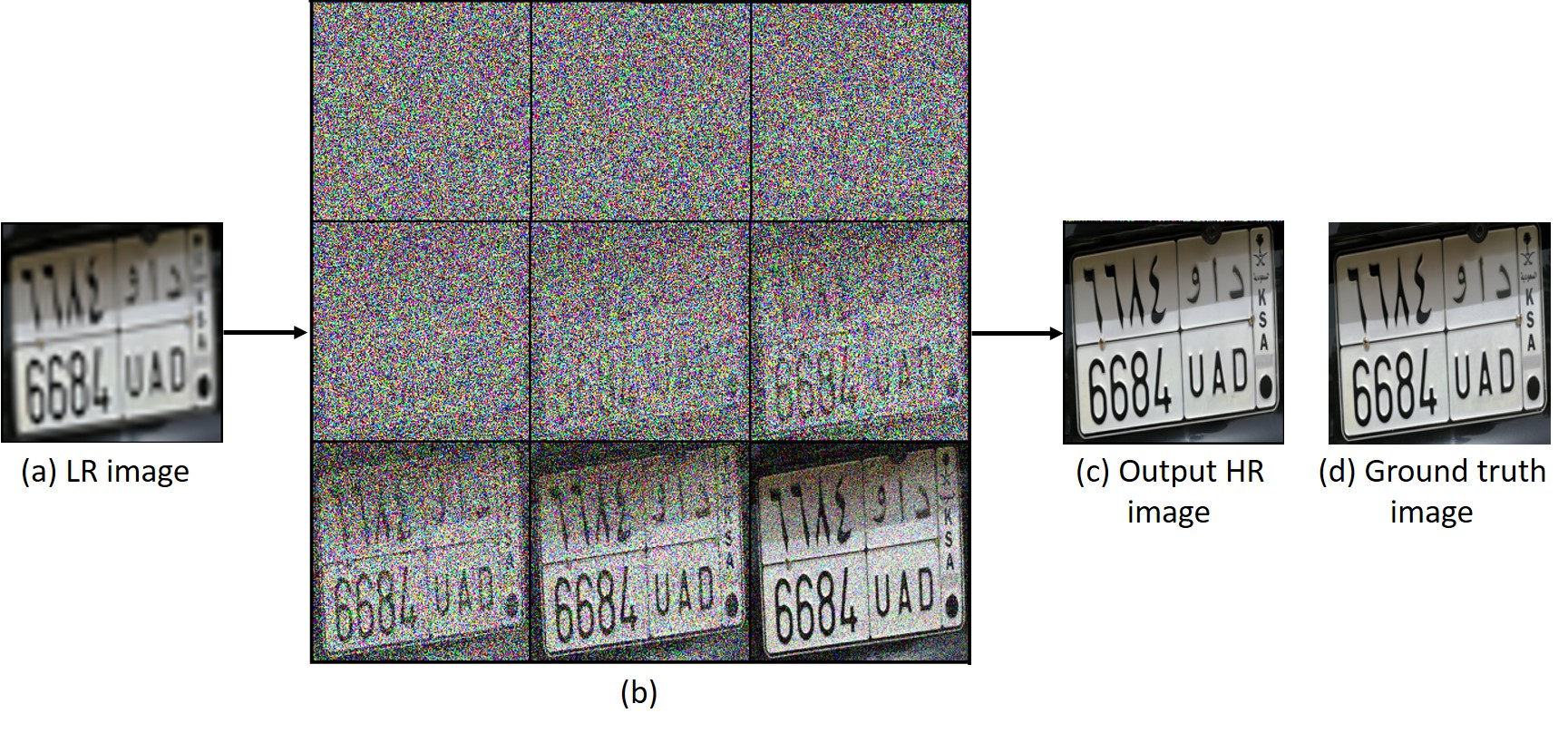}
\caption{This figure shows the intermediate steps of the HR image reconstruction. Starting from the LR image (a), then going through the denoising steps (b), to produce the output HR image (c) that resembles the ground truth image (d).  }
\label{ReconstructionSteps}
\end{figure}

Figure \ref{ReconstructionSteps} shows sample intermediate steps of the reconstruction process. It starts with the input image (a) that has \(48\times48\) resolution. With the repetitve denoising steps following the diffusion model (visualized in (b)), we were able to generate a high definition image (c) that has \(192\times192\) resolution which resembles the ground truth image (d).   By comparing the two images: the generated HR image (c) with the ground truth image (d), we notice that it is difficult to find a difference and they are almost identical. 

\subsection{Results}

\begin{figure}[h]
\centering
\includegraphics[width=0.9\columnwidth]{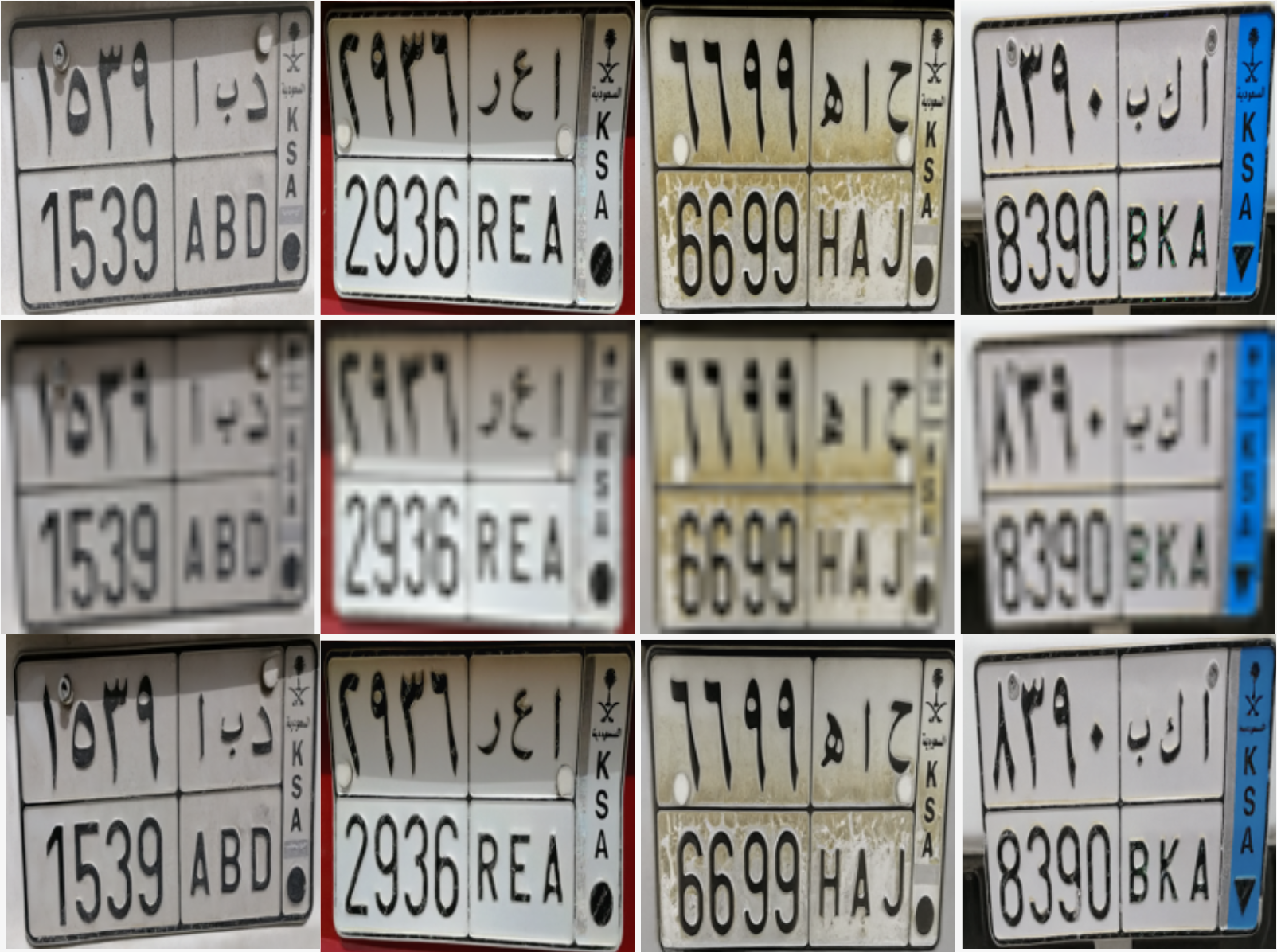}
\caption{First top row has the original HR image of (\(192\times192\)) resolution. Middle row has the downscaled LR image of (\(48\times48\)) resolution. Last row (bottom) has the super resolved images using diffusion model with (\(192\times192\)) resolution. }
\label{fig:LRtoHR}
\end{figure}

Our model was evaluated on 50 images from the LP dataset. Figure \ref{fig:LRtoHR} displays a representative high-resolution (HR) output produced using our diffusion model. These generated HR images closely match the original ground truth, preserving all image features.

\begin{figure}
     \centering
     \begin{subfigure}[b]{0.3\columnwidth}
         \centering
         \includegraphics[width=\textwidth]{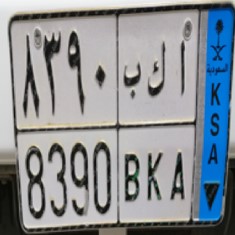}
         \caption{Ground truth}
     \end{subfigure}
     \hfill
     \begin{subfigure}[b]{0.3\columnwidth}
         \centering
         \includegraphics[width=\textwidth]{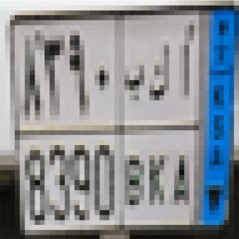}
         \caption{LR Image}
     \end{subfigure}
     \hfill
     \begin{subfigure}[b]{0.3\columnwidth}
         \centering
         \includegraphics[width=\textwidth]{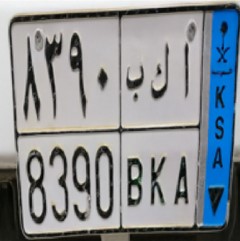}
         \caption{HR Image}
     \end{subfigure}
     \hfill
     
     
     \begin{subfigure}[b]{0.3\columnwidth}
         \centering
         \includegraphics[width=\textwidth]{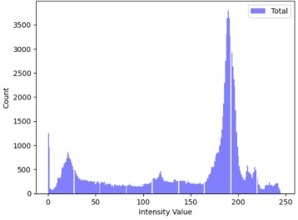}
         \caption{GT Histogram}
     \end{subfigure}
     \hfill
     \begin{subfigure}[b]{0.3\columnwidth}
         \centering
         \includegraphics[width=\textwidth]{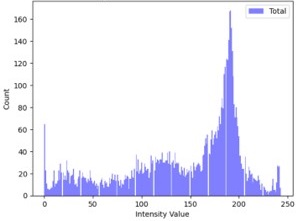}
         \caption{LR  Histogram}
     \end{subfigure}
     \hfill
     \begin{subfigure}[b]{0.3\columnwidth}
         \centering
         \includegraphics[width=\textwidth]{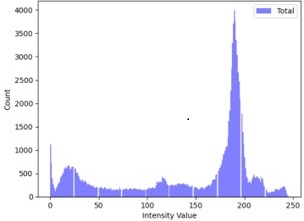}
         \caption{  HR  Histogram}
     \end{subfigure}
        \caption{First row: (a) ground truth HR image, (b) the downscaled LR image and (c) the generated HR image using diffusion model (c). The bottom row shows their respective histograms. The generated HR image histogram (f) is very similar to the ground truth image histogram (d) in contrast to the LR image histogram (e).}
        \label{fig:Histograms}
\end{figure}

To assess the fidelity of our generated images, we analyzed their histograms against the ground truth. Figure \ref{fig:Histograms} presents histograms for the original HR image, the downscaled low-resolution (LR) image, and the HR image generated by our diffusion model. While the LR image histogram markedly differs from the original due to noise, the histogram of the produced HR image is strikingly consistent with that of the original. This indicates our diffusion model's capability to reproduce HR images that retain the intrinsic characteristics of the ground truth.

\subsection{Comparison with State-of-the-art Methods}
This section discusses the comparison between our results with the results generated by two selected  state-of-the-art techniques. 
In the recent literature, there are two main approaches which are commonly used in image generation. Some techniques use encoders and decoders to synthesize images. 
Others are designed based on GAN algorithms. 
For the first approach, we selected the state-of-the-art method SwinIR \cite{SwinIR2021} while ESRGAN \cite{ESRGAN2018} was selected as the state-of-the-art  GAN-based method.  
We performed quantitative and qualitative analysis as discussed next. 

\subsubsection{Quantitative Results }
Using diffusion models to generate high resolution images in the domain of license plate images super resolution has proved to be more efficient than using the other traditional methods. We compared our approach with State-of-the-art methods, i.e. SwinIR \cite{SwinIR2021} and ESRGAN \cite{ESRGAN2018}. We used our LP images dataset and built the SwinIR and ESRGAN models for the comparison. Table \ref{table:SOTA_results} shows the quantitative comparison on the LP SR task. 
We used three metrics for the results evaluation, i.e. PSNR, SSIM and MS-SSIM, as discussed in section \ref{EvalMetrics}. The results demonstrated that our approach outperformed the SOTA methods with a significant margin using the three selected evaluation metrics.  
First, using PSNR to assess the diffusion model showed \(12.55\%\) improvement over SwinIR and  \(37.32\%\) improvement over ESRGAN.
Similarly, using SSIM, DDPM outperformed SwinIR by \(4.89\% 
\)  and ESRGAN by \(17.66\%\). 
On the other hand, MS-SSIM did not show a similar very large improvement but it  was also better. This indicates that DDPM outperforms the other traditional methods in the LP SR setting.

\begin{table} 
\caption{Using diffusion model proved its superiority to generate high resolution images in the context of license plate images. This table shows the quantitative results to compare our approach with SOTA methods: SwinIR and ESRGAN using three metric: PSNR, SSIM and MS-SSIM. }
\label{table:SOTA_results}
\centering
\setlength{\tabcolsep}{3pt}
\begin{tblr}{|p{0.09\columnwidth}|p{0.112\columnwidth}|p{0.112\columnwidth}|l|p{0.13\columnwidth}|p{0.13\columnwidth}|}
\hline
Metrics          & \textbf{Diffusion Model} & \textbf{SwinIR Transformer} & \textbf{ESRGAN} & \textbf{Improve-\\ment  Over SwinIR} & \textbf{Improve-\\ment over \\ESRGAN} \\ \hline
\textbf{PSNR}    & 30.6905                  & 24.2678                     & 22.3503         & 12.55\%                          & 37.32\%                          \\ \hline
\textbf{SSIM}    & 0.9471                   & 0.9030                      & 0.8150          & 4.89\%                           & 17.66\%                          \\ \hline
\textbf{MS-SSIM} & 0.9934                   & 0.9836                      & 0.9404          & 0.99\% $\sim$1\%              pl    & 5.64\%                           \\ \hline
\end{tblr}

\end{table}

\subsubsection{Qualitative Results: Human Evaluation}

While metrics such as PSNR and SSIM offer quantitative insights, they might not always align with human perception. To better understand the visual quality of images generated by SwinIR, ESRGAN, and our DDPM approach, we conducted a human evaluation study.

For this, we utilized the three-alternative-forced-choice (3-AFC) discrimination test, a reputable method for subjective image quality assessment, particularly when gauging an approach's effectiveness \cite{MAFC2015}. In this paradigm, participants are presented with three options and tasked with choosing the one that best meets a given criterion.

In our study, participants answered 11 questions, each featuring images produced by the aforementioned super-resolution methods. These images were randomly ordered for each participant to eliminate potential bias. The primary directive was to identify the clearest, highest-quality image from each set. A total of 50 individuals participated.

\begin{figure}[h]
\centering
\includegraphics[width=0.95\columnwidth]{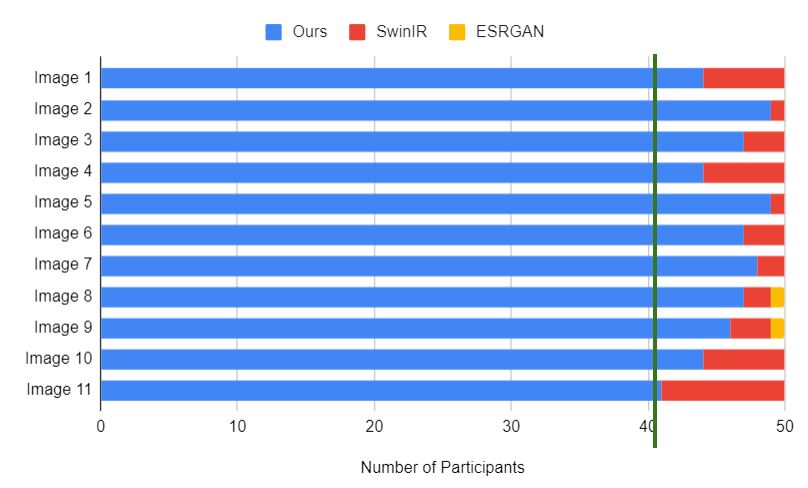}
\caption{Number of participants who selected the respective generated super resolved image as the clearest image. Blue, red and yellow colors denote the  number of participants who selected our generated image, SwinIR generated image or ESRGAN generated image, respectively. More than 40 participants among all of the 50 participants selected our generated image for all of the questions. }
\label{fig:HumanEvalHistogram}
\end{figure}

Figure \ref{fig:HumanEvalHistogram} reveals that over 40 participants consistently chose our generated images as the clearest across all questions. SwinIR was preferred by only a few, and ESRGAN was the top choice for merely two questions by a single participant. Table \ref{table:HumanEvalResults} enumerates the percentage preferences for each technique, underscoring the consistent visual superiority of our generated LP images.

\begin{table}
\caption{ Most of the participants (92\%) selected the LP images generated by our diffusion model over the other generated images as the clearest image. For the ESRGAN generated images; only one participant in two questions selected the ESRGAN generated image.} 
\label{table:HumanEvalResults}
\centering
\setlength{\tabcolsep}{3pt}
\begin{tblr}{|c|c|c|c|}
\hline
 & \textbf{SwinIR}	& \textbf{ESRGAN}	& \textbf{Diffusion Model}\\
\hline
Average percentage of \\participants who selected \\the algorithm		& 8\%			& 0\%      & 92\% \\ 
\hline
\end{tblr}

\end{table}



\subsubsection{Qualitative Results: Visual Details Evaluation}

\begin{figure}
     \centering
     \begin{subfigure}[b]{0.3\columnwidth}
         \centering
         \subcaption{SwinIR}
         \includegraphics[width=\textwidth]{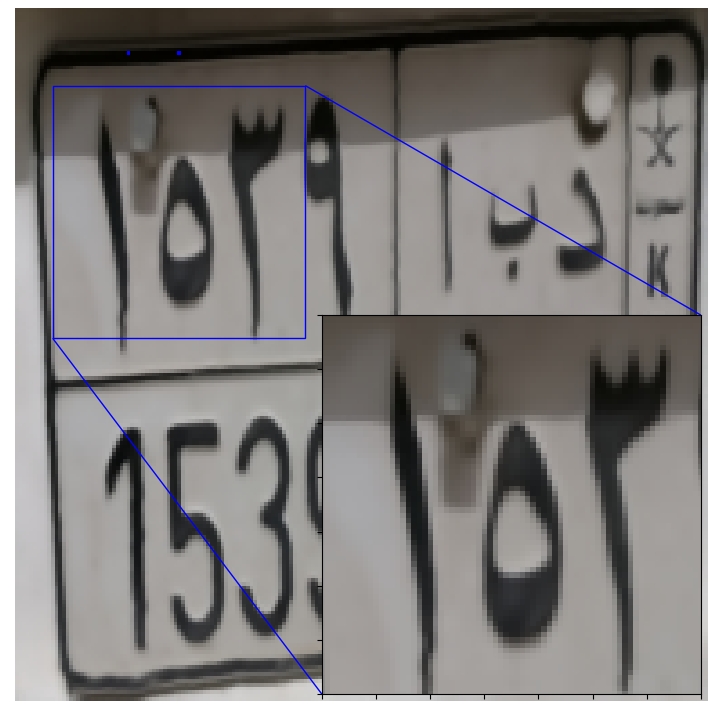}
     \end{subfigure}
     \hfill
     \begin{subfigure}[b]{0.3\columnwidth}
         \centering
         \subcaption{DDPM}
         \includegraphics[width=\textwidth]{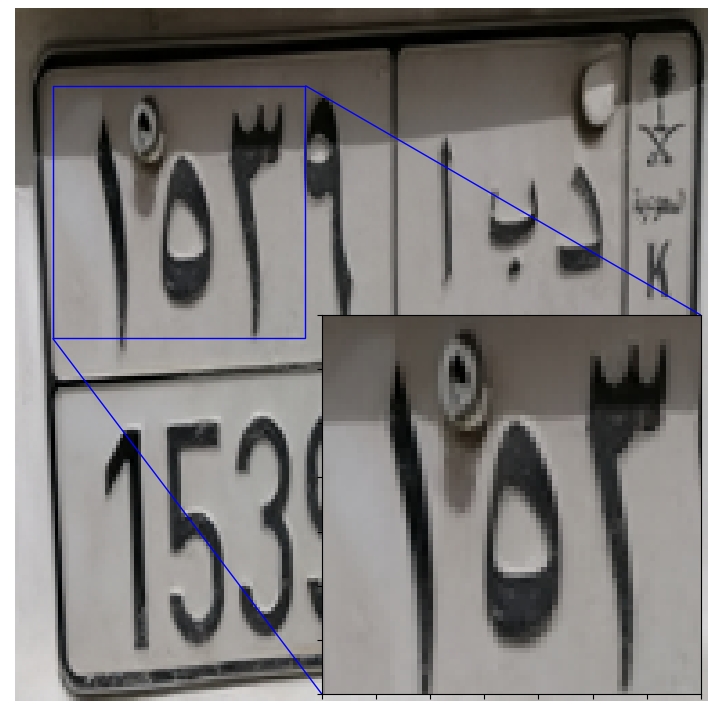}
     \end{subfigure}
     \hfill
     \begin{subfigure}[b]{0.3\columnwidth}
         \centering
         \subcaption{Original}
         \includegraphics[width=\textwidth]{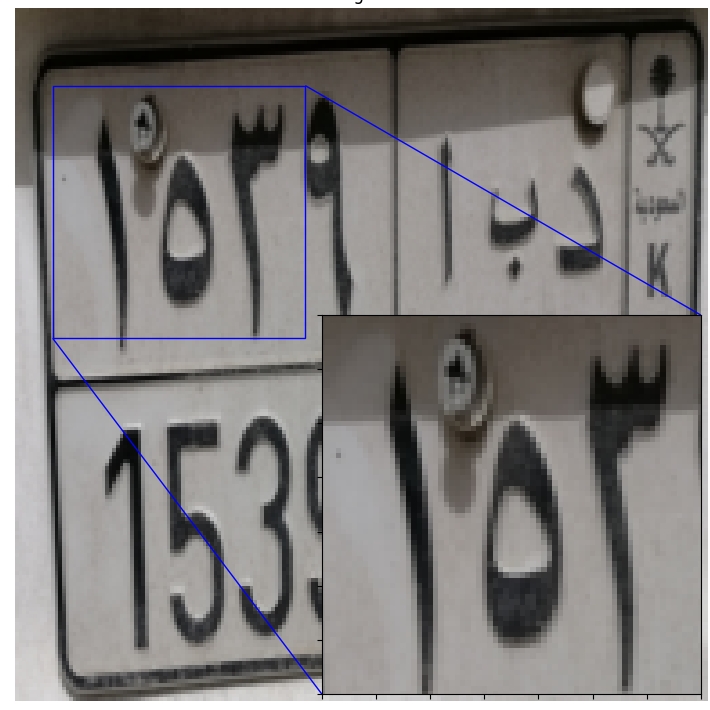}
     \end{subfigure}
     \hfill
     \\
     
     
     
     \begin{subfigure}[b]{0.3\columnwidth}
         \centering
         \includegraphics[width=\textwidth]{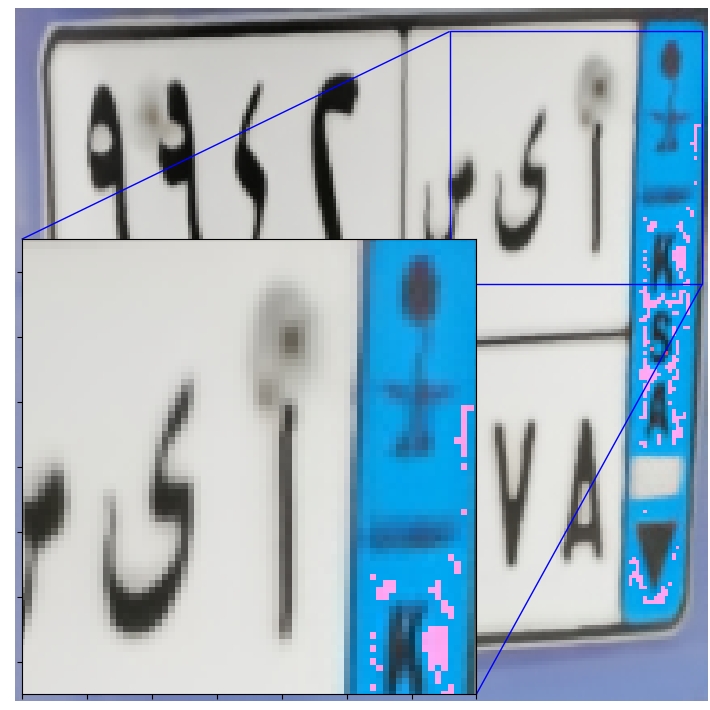}
     \end{subfigure}
     \hfill
     \begin{subfigure}[b]{0.3\columnwidth}
         \centering
         \includegraphics[width=\textwidth]{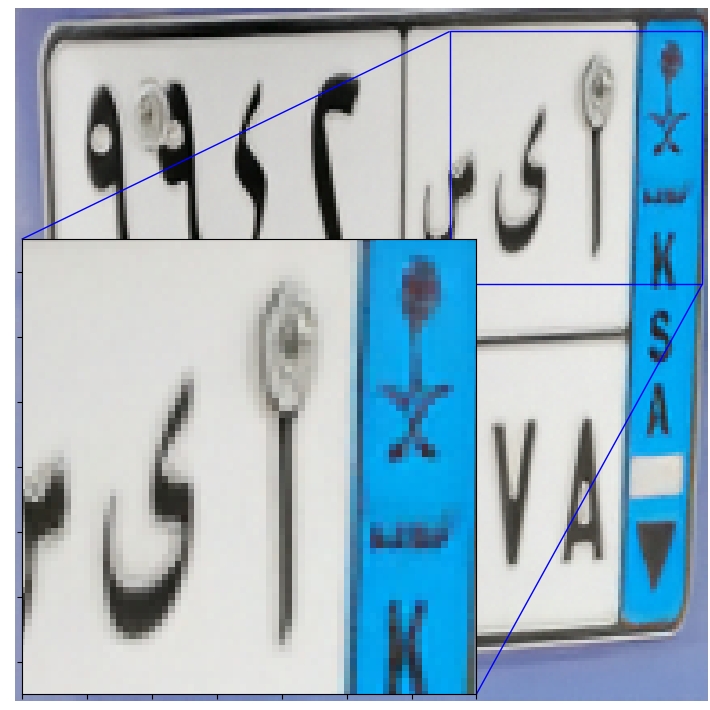}
     \end{subfigure}
     \hfill
     \begin{subfigure}[b]{0.3\columnwidth}
         \centering
         \includegraphics[width=\textwidth]{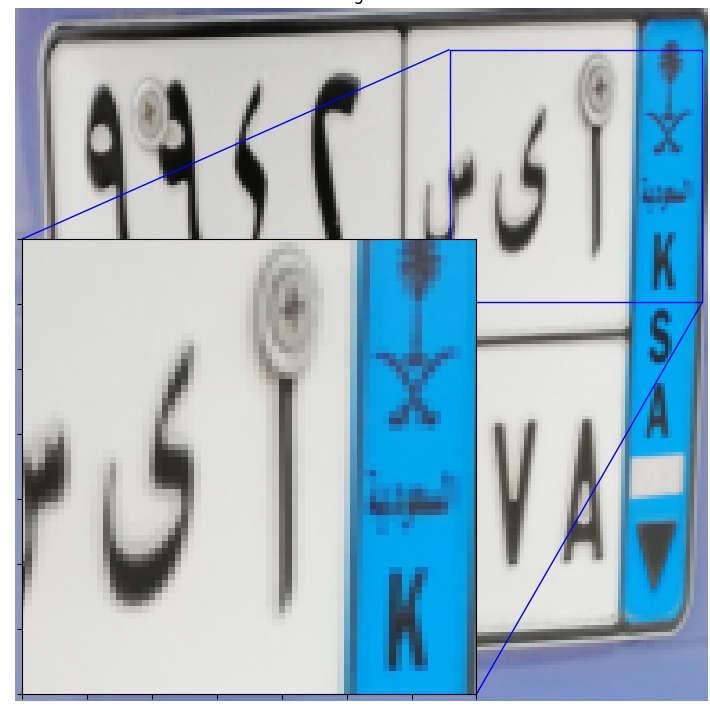}
     \end{subfigure}
        \caption{The first (left) column has the SwinIR super-resolved images, the middle column has our generated images, and the last (right) column shows the original images, with zoomed regions in each case. One can see that SwinIR failed to recover some details while our method could recover them. Our generated images are closer to the original images.  }
        \label{fig:SwinIRCompare}
\end{figure}

\begin{figure}
     \centering
     \begin{subfigure}[b]{0.3\columnwidth}
         \centering
         \subcaption{ESRGAN}
         \includegraphics[width=\textwidth]{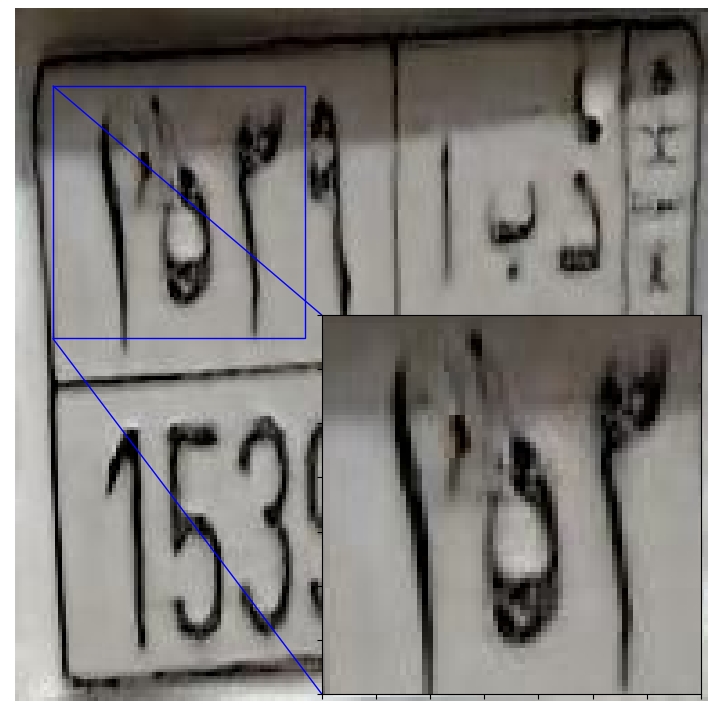}
     \end{subfigure}
     \hfill
     \begin{subfigure}[b]{0.3\columnwidth}
         \centering
         \subcaption{DDPM}
         \includegraphics[width=\textwidth]{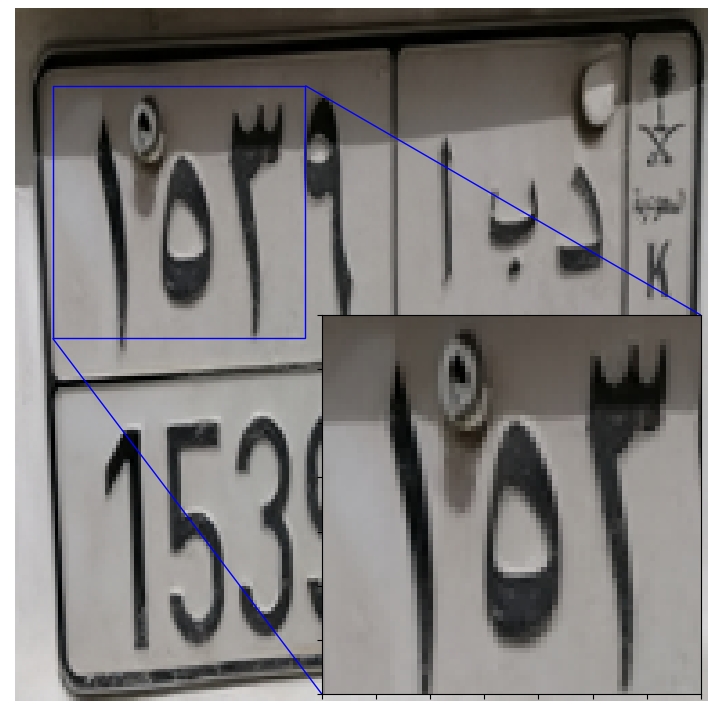}
     \end{subfigure}
     \hfill
     \begin{subfigure}[b]{0.3\columnwidth}
         \centering
         \subcaption{Original}
         \includegraphics[width=\textwidth]{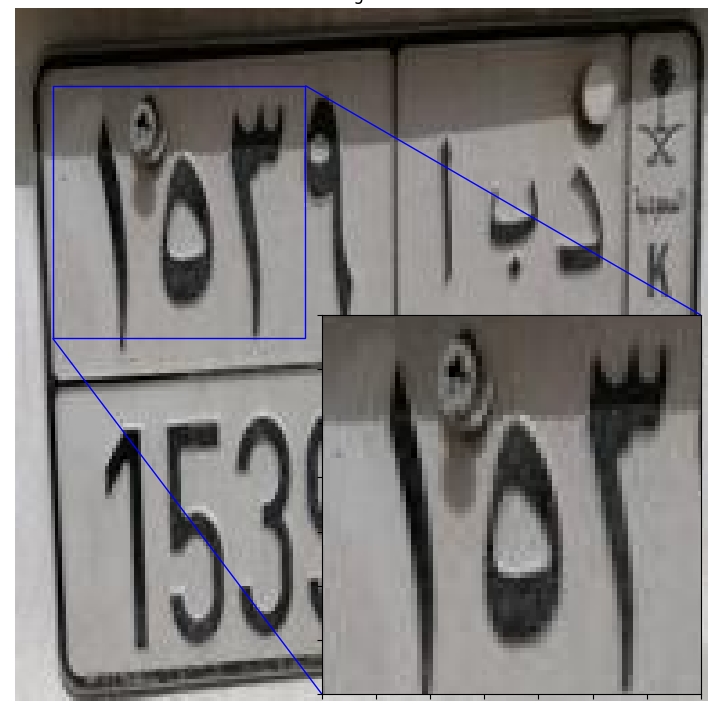}
     \end{subfigure}
     \hfill
     \\
     
     
     
     \begin{subfigure}[b]{0.3\columnwidth}
         \centering
         \includegraphics[width=\textwidth]{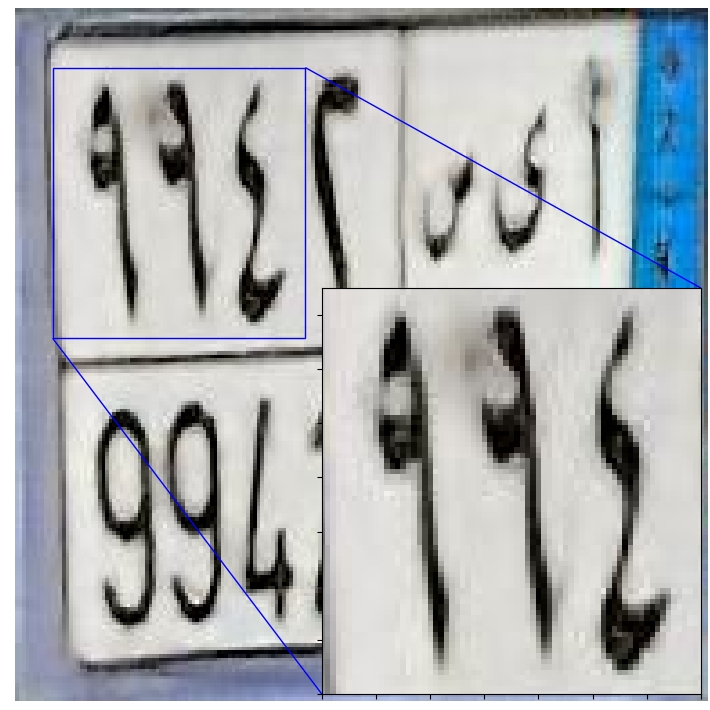}
     \end{subfigure}
     \hfill
     \begin{subfigure}[b]{0.3\columnwidth}
         \centering
         \includegraphics[width=\textwidth]{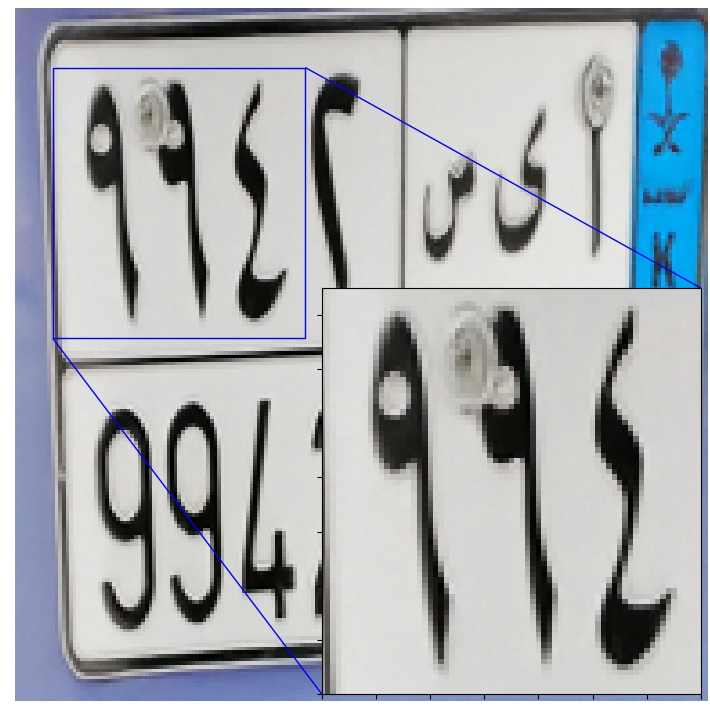}
     \end{subfigure}
     \hfill
     \begin{subfigure}[b]{0.3\columnwidth}
         \centering
         \includegraphics[width=\textwidth]{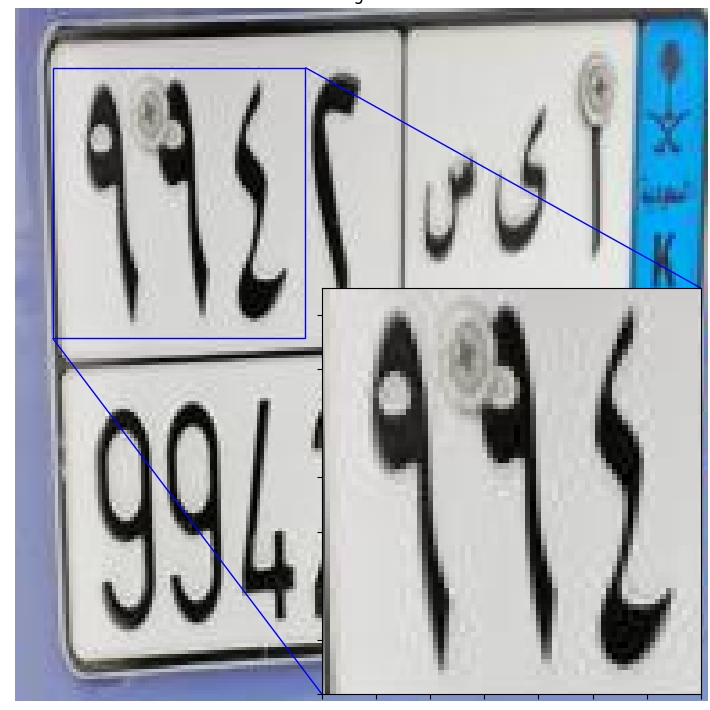}
     \end{subfigure}
        \caption{ESRGAN failed to recover many of the details in their generated images (first left column), while our generated images (middle column) have closer visual details compared to the original images (last right column).  }
        \label{fig:ESRGAN_Compare}
\end{figure}

Upon evaluating images generated by the three algorithms, our method consistently outperformed the others. Empirical results, reinforced by human assessments, highlight the preeminence of our approach. As depicted in Figure \ref{fig:SwinIRCompare}, our diffusion model demonstrates superior detail restoration capabilities compared to SwinIR. Analogously, Figure \ref{fig:ESRGAN_Compare} showcases our model's enhanced prowess in rendering intricate details of super-resolved LP images. Particularly noteworthy is the pronounced sharpness of our images, especially around character boundaries.

\section{Conclusions} \label{Conclusions}
In this paper, we presented a generative model that would restore a high quality image from a highly distorted image. To the best of our knowledge, we believe that we are the first to utilize diffusion models in the context of license plate images super resolution. Our results showed that our proposed approach has superior performance compared to other traditional methods.Using diffusion model for LP super resolution significantly outperforms SOTA methods such as SwinIR and ESRGAN with a notable margin. Our model had \(13\%\) PSNR improvement over SwinIR and \(37\%\)  over ESRGAN. 
Similarly, the score of SSIM was \(5\%\) better than SwinIR and \(18\%\) better than ESRGAN. 
Besides, our model was able to capture detailed features in the LP images which, on the other side, failed by the other SOTA methods. 
According to our human evaluation experiment, our participants selected our generated images 92\% as the clearest image, against SwinIR and ESRGAN images. 
This shows that diffusion model is a good candidate for improving the quality of license plate images, hence, improving the performance of LP recognition systems. 
On the other hand, the main disadvantage of using the diffusion model approach is its heavy computational cost. This hinders it use especially in real time applications. However and by considering the superior performance of diffusion model, it is worth considering this approach in super resolution problems. In the future, we will work in the direction of minimizing its computational costs while harnessing its powerful performance.


\appendices

\section*{Acknowledgment}
 The authors thank Prince Sultan University for their support and funding in conducting this research.

\bibliographystyle{IEEEtran}
\bibliography{LP_Diffusion}

\end{document}